\documentclass[a4paper,10pt,twocolumn]{article}

\usepackage[english]{babel}
\usepackage[utf8]{inputenc}
\usepackage[T1]{fontenc}

\usepackage[top=1.5cm, left=1.5cm, right=1.5cm, bottom=1.5cm]{geometry}

\renewenvironment{abstract}{\bf\small {\em\ Abstract---}}{}

\usepackage{amsfonts,amssymb,amsmath,amsthm}
\usepackage{subfigure}
\usepackage{graphicx}
\usepackage[footnotesize]{caption}

\usepackage{amssymb} 
\usepackage{amsfonts}
\usepackage{amsmath} 
\usepackage{cite}
\usepackage{bm} 
\usepackage{graphicx} 
\usepackage{cite} 
\usepackage{algorithmic}
\usepackage{algorithm}
\usepackage{float}
\usepackage{url}
\usepackage{mathtools}
\usepackage{multirow}
\usepackage{diagbox}
\usepackage{booktabs}
\usepackage{threeparttable}

\def\defn{\,\triangleq\,} 
\def\d{\, \mathrm{d}} 
\def\im{\mathrm{j}} 

\def\uin{{u_{\text{\tiny in}}}}
\def\usc{{u_{\text{\tiny sc}}}}

\def\Real{\textsf{Re}}
\def\Imag{\textsf{Im}}

\newcolumntype{L}[1]{>{\raggedright\arraybackslash}p{#1}}
\newcolumntype{C}[1]{>{\centering\arraybackslash}p{#1}}
\newcolumntype{R}[1]{>{\raggedleft\arraybackslash}p{#1}}

\def\wbf{\mathbf{w}}

\def\ubf{\mathbf{u}}
\def\ubfin{{\mathbf{u}_{\text{\tiny in}}}}
\def\ubfinast{{\mathbf{u}_{\text{\tiny in}}^\ast}}
\def\xbf{\mathbf{x}}

\def\ybf{\mathbf{y}}

\def\ebf{\mathbf{e}}

\def\Sbf{\mathbf{S}}

\def\Gbf{\mathbf{G}}
\def\Pbf{\mathbf{P}}

\def\rbm{\bm{r}}
\def\sbm{\bm{s}}

\def\rbmp{{\bm{r}^\prime}}

\def\kbm{\bm{k}}
\def\fbm{\bm{f}}

\def\C{\mathbb{C}}
\def\R{\mathbb{R}}

\def\diag{\textsf{diag}}

\def\Hrm{\textsf{H}}

\usepackage{xspace}
\usepackage{xcolor}

\title{Stability of Scattering Decoder for Nonlinear Diffractive Imaging}

\author{Yu Sun$^1$ and Ulugbek S. Kamilov$^{1,2}$\\
  \footnotesize $^1$ Department of Computer Science \& Engineering, Washington University in St Louis. \\
  \footnotesize $^2$ Department of Electrical \& Systems Engineering, Washington University in St. Louis } \date{\empty} 

\begin{document}

\maketitle

\begin{abstract}
The problem of image reconstruction under multiple light scattering is usually formulated as a regularized non-convex optimization. A deep learning architecture, Scattering Decoder (ScaDec), was recently proposed in \cite{Sun:18} to solve this problem in a purely data-driven fashion. The proposed method was shown to substantially outperform optimization-based baselines and achieve state-of-the-art results. In this paper, we thoroughly test the robustness of ScaDec to different permittivity contrasts, number of transmissions, and input signal-to-noise ratios. The results on high-fidelity simulated datasets show that the performance of ScaDec is stable in different settings.
\end{abstract}

\section{Introduction}
\label{sec:introduction}
The problem of reconstructing the spatial distribution of the dielectric permittivity of an unknown object by measuring the corresponding scattered light field is fundamental in many applications such as optical diffractive tomography \cite{Lim.etal2015} and digital holography~\cite{Brady.etal2009}. Consider an object with the permittivity distribution $\epsilon(\rbm)$ is centered in a bounded domain ${\Omega \subseteq \R^2}$, with a background medium of permittivity $\epsilon_b$. The object is illuminated by a monochromatic and coherent incident electric field $\uin(\rbm)$, which is assumed to be known both inside $\Omega$ and at the sensor domain ${\Gamma \subseteq \R^2}$. The light field $\usc(\rbm)$ scattered by the object is collected at $\Gamma$ as measurements. The interaction between the object and the wave can be mathematically described by the Lippmann-Schwinger equation \cite{Born.Wolf2003} 
\begin{equation}
\label{Eq:ImageField}
u(\rbm) = \uin(\rbm) + \int_{\Omega} g(\rbm - \rbmp) \, f(\rbmp) \, u(\rbmp) \d \rbmp,\quad(\rbm \in \R^{2})
\end{equation}
where $u(\rbm) = \uin(\rbm) + \usc(\rbm)$ is the total light field. The scattering potential, assumed to be real, is defined as ${f(\rbm) \defn k^2 (\epsilon(\rbm)-\epsilon_b)}$, where $k = 2\pi/\lambda$ is the wavenumber. The Green's function $g(\rbm)$ for two-dimensional free space is defined as $g(\rbm) \defn \frac{\im}{4} H_0^{(1)}(k_b \|\rbm\|_{\ell_2})$, where $H_0^{(1)}$ denotes the zero-order Hankel function of the first kind and $k_b \defn k \sqrt{\epsilon_b}$ is the wavenumber of the background medium. The discrete system that models wave-object interaction is given by
\begin{subequations}
\label{Eq:NonlinearModel}
\begin{align}
&\ubf = \ubfin + \Gbf (\ubf \odot \xbf) \label{Eq:ForwardScat2} \\
&\ybf = \Sbf(\ubf \odot \xbf) + \ebf \label{Eq:ForwardScat1}\;, 
\end{align}
\end{subequations}
where $\xbf \in \R^N$ is the discretized scattering potential $f$ of the object, $\ybf \in \C^M$ is the measured scattered field $\usc$ at $\Gamma$, ${\ubfin \in \C^N}$ is the input field $\uin$ inside $\Omega$, $\Sbf \in \C^{M \times N}$ and ${\Gbf \in \C^{N \times N}}$ are the discretizations of the Green's functions in (\ref{Eq:ImageField}) evaluated inside $\Omega$ and at $\Gamma$, respectively, $\odot$ denotes a component-wise multiplication between two vectors, and $\ebf \in \C^M$ models the additive noise at the measurements.
\begin{figure}[t]
\begin{center}
\includegraphics[width=7.8cm]{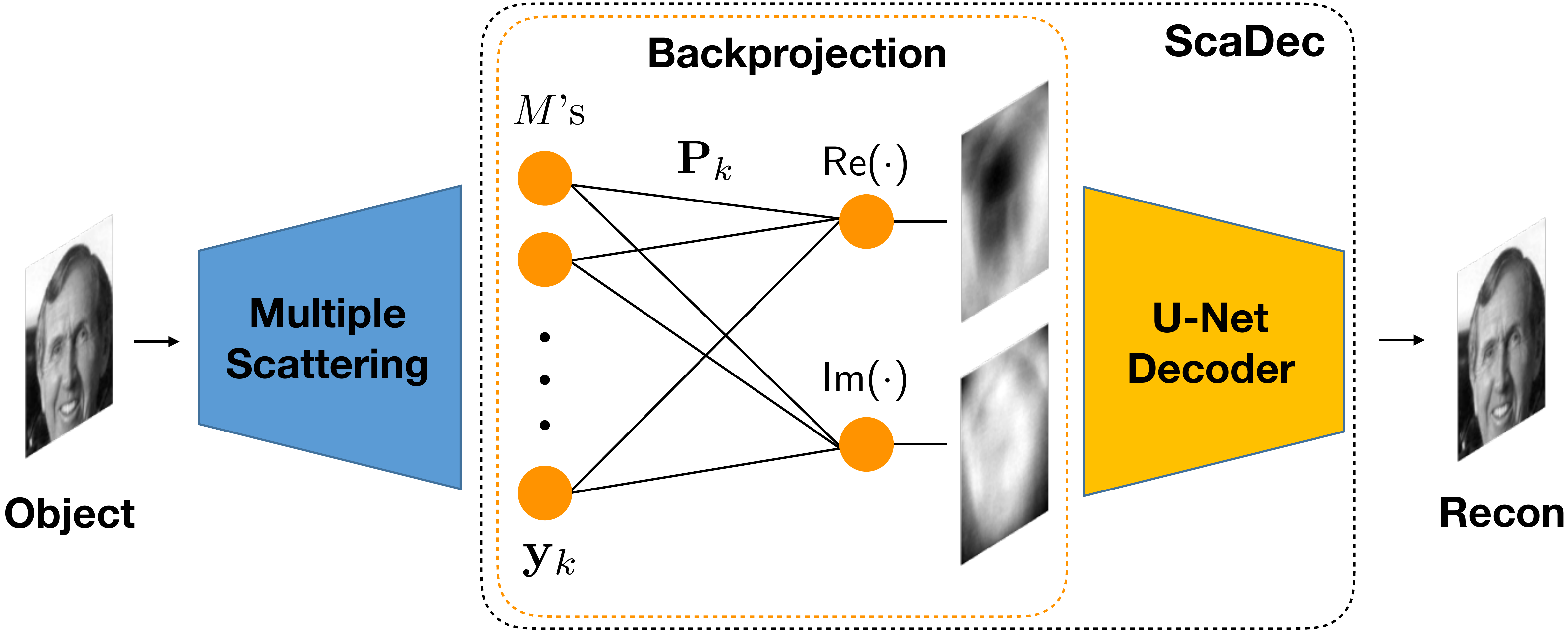}
\end{center}
\caption{ScaDec in \cite{Sun:18} consists of backprojection from measurements to a complex data followed by a ConvNet mapping the data to the final image.}
\label{Fig:ScaDec}
\end{figure}
Commonly, the problem of image reconstruction under multiple light scattering is formulated as a non-convex optimization, where a nonlinear forward model is used to simulate multiple light scattering and regularization is applied to promote the restoration quality~\cite{Tian.Waller2015, Kamilov.etal2015b, Kamilov.etal2016, Kamilov.etal2016a, Soubies.etal2017, Liu.etal2018, Pham.etal2018, Maire.etal09, Arhab.etal13}. Different from optimization-based methods, the recent paper \cite{Sun:18} proposed a novel deep learning model, called Scattering Decoder (ScaDec), to reconstruct image under multiple scattering. By interpreting multiple scattering as a forward pass of a convolutional neural network (ConvNet), \cite{Sun:18} considered to reconstruct image by designing a deep ConvNet to invert multiple scattering in a purely data-driven fashion. 

Figure~\ref{Fig:ScaDec} illustrates the general framework of ScaDec. The first component in the model simply backprojects the measurement data to the image domain. The mathematical expression of backprojection is specified by
\begin{equation}
\label{Eq:BackPropagationMultiple}
\wbf = \sum_{k = 1}^K{\Pbf_{k} \ybf_{k}} \;, \;\text{with}\; \Pbf_{k} \defn \diag(\ubfinast_{,k}) \Sbf^\Hrm\\
\end{equation}
where vector $\ybf_{k} \in \C^{M}$ are the measurements of the $k$th transmission and collected by $M$ receivers, and matrix $\Pbf_{k} \in \C^{N \times M}$ is the backprojection operator. Inside the operator, matrix $\Sbf^\Hrm \in \C^{N \times M}$ is the Hermitian transpose of the discretized Green's function $\Sbf$, and $\ubfinast_{,k}$ is the element-wise conjugate of the incident wave field of the $k$th transmission. The output $\wbf \in \C^N$ is the summation of the projected images of $K$ transmissions. The backprojection also can be viewed as a fixed layer in a ConvNet with $\Pbf_{k}$ characterizing the weights, and $\Real(\cdot)$ and $\Imag(\cdot)$ featuring the activation functions, respectively. 

The second component is a convolutional neural network based on the U-Net architecture~\cite{RFB15a, Jin.etal2016, Ye.etal2018,Borhani:18}. Jointly with the backprojection, the U-Net decoder builds an end-to-end mapping from measurements of multiple scattered wave to the spatial distribution of the permittivity of the object. Comparison on simulated and experimental datasets in \cite{Sun:18} shows that ScaDec significantly outperforms other optimization-based baselines in terms of both reconstruction quality and time complexity. In this paper, we further evaluate the stability of ScaDec in the scenarios of different permittivity contrasts, number of transmissions, and input signal-to-noise ratios (SNR) on high-fidelity simulated datasets. The experimental results, to be shown in next section, concur that the performance of ScaDec is stable in various situations.

\section{Main Result}
\label{sec:first-section}
We now test the stability of ScaDec with respect to variations in three individual aspects: 1) permittivity contrast, 2) number of transmissions, and 3) input SNR. In the experiments, we used the dataset of human faces \cite{liu2015faceattributes}, and the measurements were obtained by solving the Lippmann-Schwinger equations with a conjugate-gradient solver \cite{Liu.etal2018}. The dataset contains 1500 images for training, 24 for validating, and 24 for testing which were randomly selected from untouched images.

\begin{figure}[t]
\begin{center}
\includegraphics[width=7.5cm]{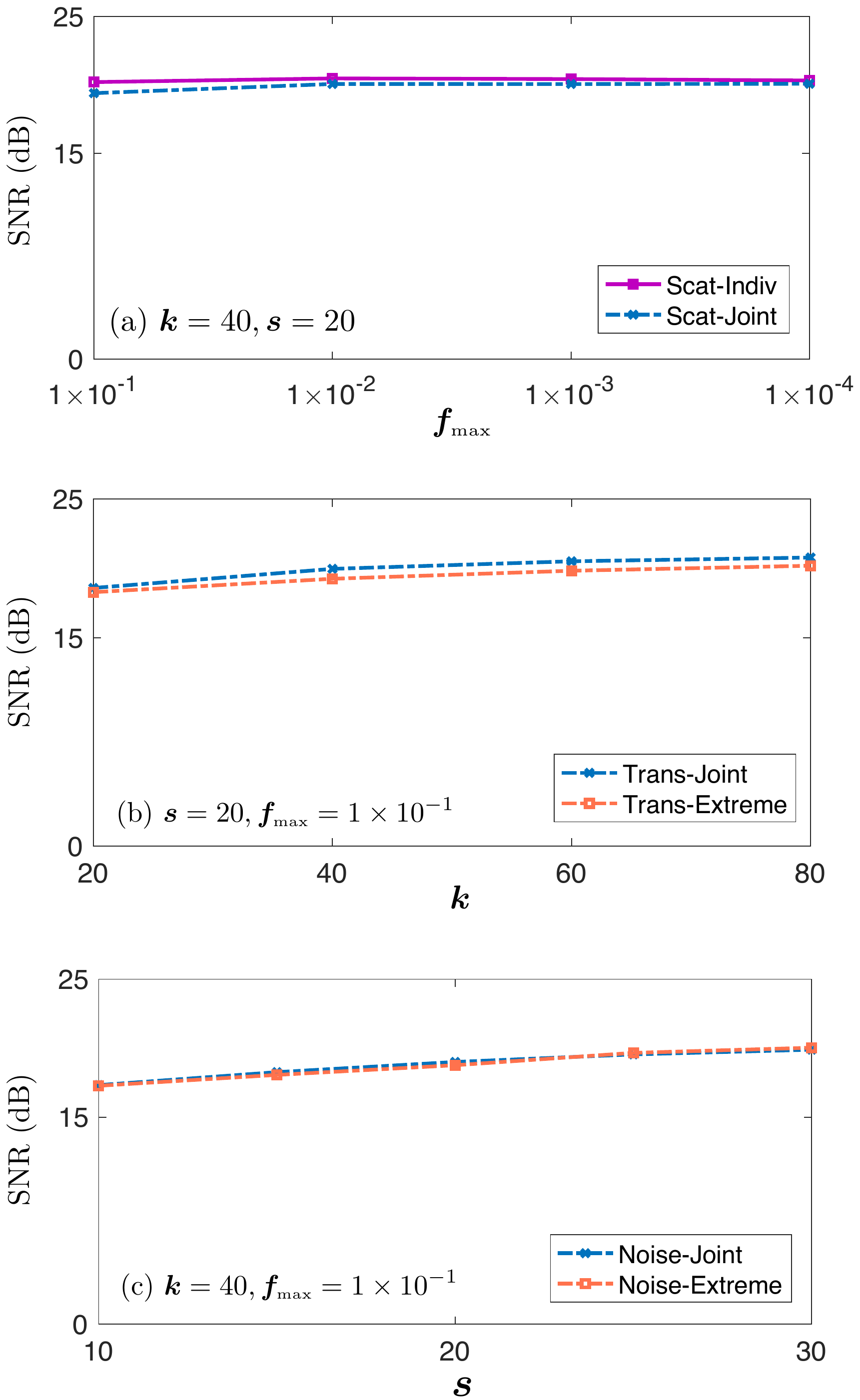}
\end{center}
\caption{The performance of ScaDec with respect to (a) permittivity contrast $\fbm_\text{\tiny max}$, (b) transmittion number $\kbm$, and (c) input SNR $\sbm$. The corresponding parametric defaults are presented in the bottom left corner.}
\label{Fig:RobustTest}
\end{figure}

The physical size of images was set to 18 cm $\times$ 18 cm, discretized to a $128 \times 128$ grid. We define the permittivity contrast as $\fbm_\text{\tiny max} \defn (\epsilon_{\text{\tiny max}} - \epsilon_b)/\epsilon_b$, where $\epsilon_{\text{\tiny max}} \defn \max_{\rbm \in \Omega} \{\epsilon(\rbm)\}$. The background medium was assumed to be air with $\epsilon_b = 1$ and the wavelength of the illumination was set to $\lambda = 0.84$ cm. Total $\kbm$ transmissions were uniformly distributed along a circle of radius $1.6$ m and for each transmission 360 measurements were collected around the image. The simulated measured data was further corrupted by an additive Gaussian white noise consistent with $\sbm$ dB of input SNR. The noise model is mathematically modeled by the $\ell_2$-norm, which is common in the loss function for training a neural network.

\begin{figure}[t]
\begin{center}
\includegraphics[width=7.6cm]{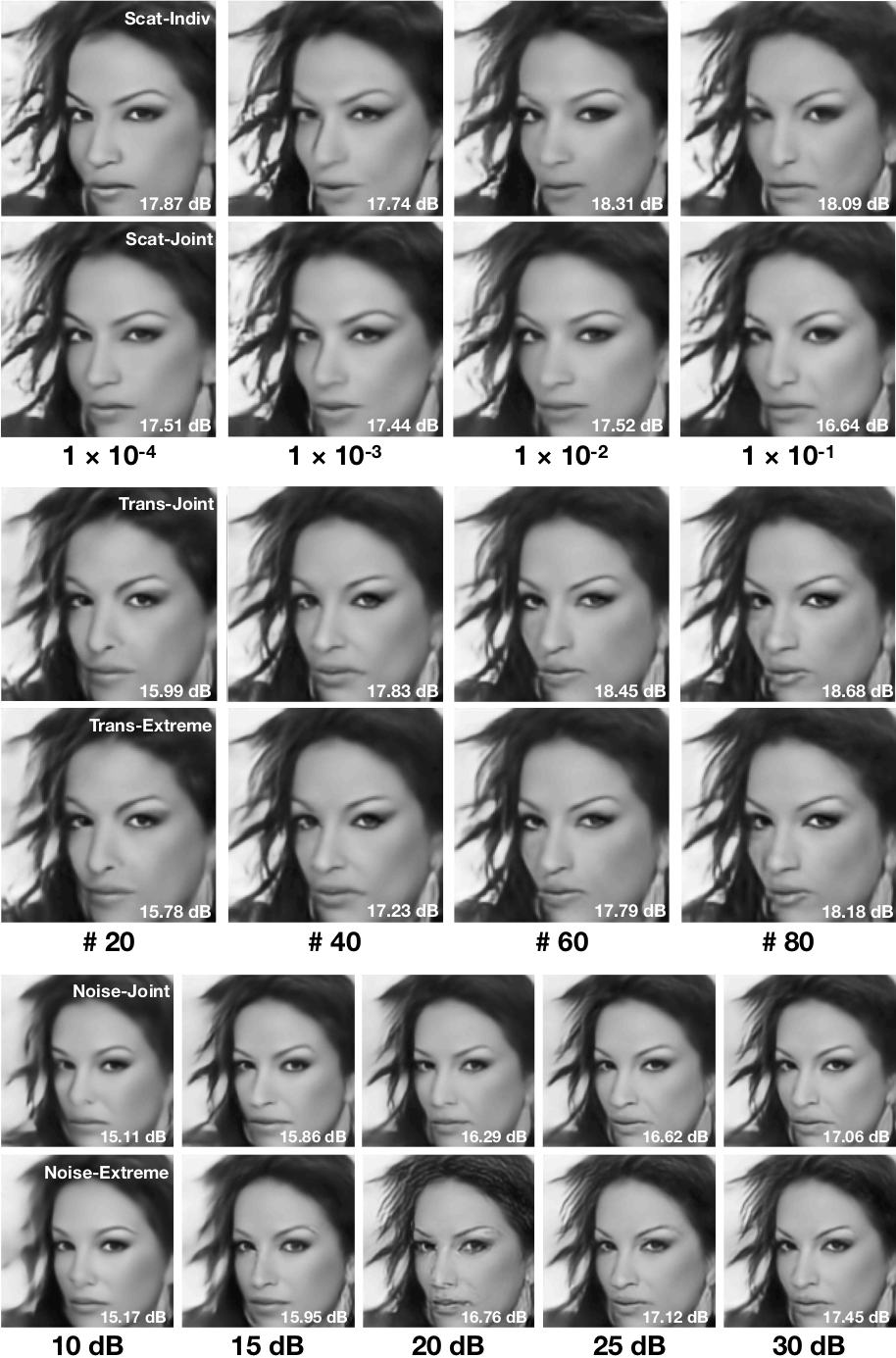}
\end{center}
\caption{Visual examples obtained in the three experiments. The top two rows corresponds to Scat-Indiv and Scat-Joint, middle two rows to Trans-Joint and Trans-Extreme, bottom two rows to Noise-Joint and Noise-Extreme.}
\label{Fig:VisualExample}
\end{figure}

Fig.~\ref{Fig:RobustTest} and Fig.~\ref{Fig:VisualExample} empirically and visually evaluate the robustness of ScaDec regarding to different contrasts $\fbm_\text{\tiny max}$, numbers of illuminations $\kbm$, and input SNR $\sbm$. The parametric setting of each test is reported in the corresponding plot. Fig.~\ref{Fig:RobustTest}(a) summarizes the performance of ScaDec with respect to different $\fbm_\text{\tiny max}$. Scat-Indiv estimated the optimal performance by individually training on the data with $\fbm_\text{\tiny max}$ equals to $1\times10^{-1}$, $1\times10^{-2}$, $1\times10^{-3}$ and $1\times10^{-4}$, though Scat-Joint was jointly trained on the data corresponding to all levels of $\fbm_\text{\tiny max}$. The jointly trained ScaDec obtains nearly optimal performance in the sense that the reconstruction SNR of Scat-Joint agrees with that of Scat-Indiv. 

Fig.~\ref{Fig:RobustTest}(b) and~\ref{Fig:RobustTest}(c) illustrate the stability of ScaDec as $\kbm$ and $\sbm$ vary. The blue lines correspond to the model jointly trained on the data of all the values (marks in the curves), and the orange lines represent the model trained merely on the boundary values (eg. $\kbm=10,80$). Both plots clearly show that ScaDec is relatively stable since the degradation of reconstruction SNR is gradual as $\kbm$ and $\sbm$ shift from the lower limit to the upper limit. For example, the SNR values were 18.29 dB, 19.26 dB, 19.84 dB, and 20.20 dB at $\kbm$ equal to 20, 40, 60, and 80, respectively. Moreover, ScaDec shows good ability to generalize since the extreme models matches the joint models at each level of $\kbm$ and $\sbm$. It is worth of mentioning that data of the in-between levels remained unused by the former models while used by the later ones in the training.

We further explored the performance of ScaDec under extreme conditions. We considered two scenarios: 1) reconstruction from very noisy measurements ($\sbm=5$) where common algorithms fail and 2) reconstruction of images with mismatched size ($256 \times 256$) whose measurements is nonlinear to the training ones. ScaDec was reported to fail in both scenarios since the measurements are too corrupted to extract useful information and the nonlinear relationship is not easy to generalize. However, ScaDec is stable in the various and further show its potential for applications of diffractive imaging.



\bibliographystyle{ieeetr}

\end{document}